\documentclass[sigconf]{acmart}

\usepackage{booktabs} 
\usepackage{balance}

\graphicspath{ {images/} }

\begin{document}
\title{Sync-DRAW: Automatic Video Generation using Deep Recurrent Attentive Architectures}

\author{Gaurav Mittal}
\authornote{Equal Contribution}
\affiliation{
	\institution{}
    }
\email{gaurav.mittal.191013@gmail.com}

\author{Tanya Marwah{\normalsize$^{^*}$}}
\affiliation{
  \institution{IIT Hyderabad}
}
\email{ee13b1044@iith.ac.in}

\author{Vineeth N Balasubramanian}
\affiliation{
  \institution{IIT Hyderabad}
}
\email{vineethnb@iith.ac.in}


\copyrightyear{2017}
\acmYear{2017}
\setcopyright{acmcopyright}
\acmConference{MM'17}{}{October 23--27, 2017, Mountain View, CA, USA.} \acmPrice{15.00} \acmDOI{http://dx.doi.org/10.1145/3123266.3123309}
\acmISBN{ISBN 978-1-4503-4906-2/17/10}


\begin{abstract}
This paper introduces a novel approach for generating videos called Synchronized Deep Recurrent Attentive Writer (Sync-DRAW). Sync-DRAW can also perform text-to-video generation which, to the best of our knowledge, makes it the first approach of its kind. It combines a Variational Autoencoder~(VAE) with a Recurrent Attention Mechanism in a novel manner to create a temporally dependent sequence of frames that are gradually formed over time. The recurrent attention mechanism in Sync-DRAW attends to each individual frame of the video in sychronization, while the VAE learns a latent distribution for the entire video at the global level. Our experiments with Bouncing MNIST, KTH and UCF-101 suggest that Sync-DRAW is efficient in learning the spatial and temporal information of the videos and generates frames with high structural integrity, and can generate videos from simple captions on these datasets.\footnote{Accepted in ACM Multimedia 2017 to be held at Mountain View, CA, USA.}
\end{abstract}




\maketitle

\section{Introduction}
Over the years, several generative methods have been proposed to capture and model the latent representations of high-dimensional data such as images and documents. In recent years, with the success of deep learning methods, Variational Auto-Encoders (VAEs)~\cite{kingma} and Generative Adversarial Networks (GANs)\cite{goodfellow} have emerged to be the most promising deep learning-based generative methods for performing tasks such as image generation~\cite{draw,dcgan}.

Existing efforts (discussed in Section \ref{relatedwork}) in the last couple of years have primarily focused on generation of images using the aforementioned methods, by modeling the spatial characteristics of a given training dataset \cite{draw,aligndraw,van2016pixel}, as well as extending these methods to more novel applications such as adding texture/style to images \cite{gatys2015neural}. However, there has been very little effort in the automated generation of videos. While there have been a few efforts in video prediction~\cite{nitish_icml2015}, the recent work by Vondrick et al.~\cite{vondrick2016generating} was the first to generate videos in an unsupervised manner using GANs. Despite acceptable performance, the generated videos had several drawbacks, as highlighted later in this work. Our work is an attempt to provide a different perspective to video generation using VAEs, which overcomes a few drawbacks of ~\cite{vondrick2016generating}; and more importantly, provides a methodology to generate videos from captions for the first time. 

\begin{figure}
\centering
\includegraphics[width=1\columnwidth]{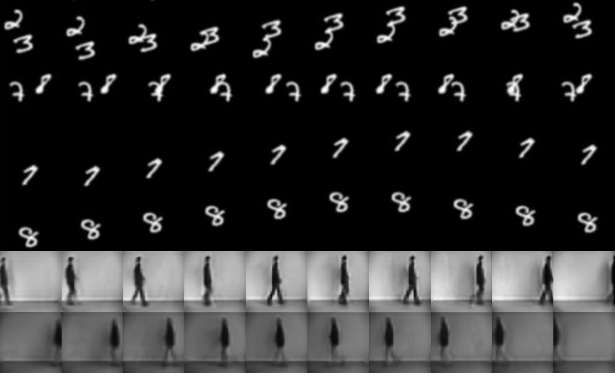}
\vspace{-20pt}
\caption{Videos generated automatically by proposed Sync-DRAW for Single-Digit Bouncing MNIST, Two-Digit Bouncing MNIST, and KTH Dataset.}
\vspace{-20pt}
\label{collage}
\end{figure}

We propose a novel network called Sync-DRAW, which uses a recurrent VAE to automatically generate videos. It utilizes a local attention mechanism which attends to each frame individually in \textit{`synchronization'} and hence, the name \textit{Sync-DRAW (Synchronized Deep Recurrent Attentive Writer)}. Our experiments show that the proposed attention mechanism in Sync-DRAW plays a significant role in generating videos that maintain the structure of objects to a large extent. Sync-DRAW also takes a step further by learning to associate videos with captions, and thus learning to generate videos from just text captions. To the best of our knowledge, this work is the first effort to generate video sequences from captions, and is also the first to generate videos using VAEs. Figure~\ref{collage} shows examples of video frames generated by trained Sync-DRAW models on different datasets. 

The key contributions of this paper lie in: (i) the design of an efficient and effective deep recurrent attentive architecture to generate videos that preserves the structural integrity of objects in each frame; (ii) the adaptation of the proposed architecture to generate videos from just text captions; and (iii) a comprehensive review and experimental study of video generations on datasets of varying complexity. We now discuss earlier efforts related to this work.

\section{Background and Related Work}
\label{relatedwork}
In order to effectively generate meaningful structured data such as an image, a learning model should be capable of learning the latent representations for the given data, thus explaining the success of deep architectures \cite{bengio}, which are today synonymous with representation learning, as generative models. Initial efforts in generating such data using deep architectures were based on Deep Belief Networks~\cite{hinton2006fast} and Deep Boltzmann Machines~\cite{salakhutdinov2009deep}, which contained many layers of latent variables and millions of parameters. However, their overwhelming dependence on Markov Chain Monte Carlo-based sampling methods made them difficult to scale. Over the last few years, newer deep learning approaches such as Generative Adversarial Networks (GANs)~\cite{goodfellow} and Variational Auto-Encoders (VAEs)~\cite{kingma} have been proven to perform quite effectively on generative tasks, and have been shown to scale to large datasets.

The Deep Recurrent Attentive Writer~(DRAW)~\cite{draw} was the earliest work to utilize a Recurrent-VAE (R-VAE) to learn to generate images by progressively refining an image canvas over time using a sequence of samples generated from a learnt distribution. DRAW further combines this R-VAE with an attention mechanism~\cite{larochelle2010learning}, which allows the generated image to focus on certain parts of the image at each timestep. This notion of attention resembles human visual perception, and has been used recently for related applications such as generating captions for images~\cite{showatell} or even generating images from captions~\cite{aligndraw,reed2016generative}. However, spatial attention differs from spatiotemporal attention, and there has been very little effort on using spatio-temporal latent representations to generate image sequences or videos, which we address in this work. Further, to the best of our knowledge, there has been no prior work in text-to-video generation, which is one of the key objectives of this work.

The state-of-the-art in the domain of video generation is a recently proposed method called VGAN by Vondrick et al. \cite{vondrick2016generating}, which attempts to generate videos using a spatio-temporal GAN. This GAN employs a two-stream strategy with one stream focusing on the foreground, and another on the static background. While the method delivers promising results, closer observation shows that its generations are fairly noisy as well as lack `objectness' (the structural integrity of the object), as shown in Section \ref{sec_vgan_compare}. In this work, we propose a new and different perspective to video generation based on Recurrent VAEs (instead of GANs), which uses a novel attention mechanism to generate videos. We show in this work that the videos generated by the proposed approach are perceived well by human users, as well as maintain the objects' structural integrity to a significant extent on visual inspection. Moreover, we extend the R-VAE to perform text-to-video generation, which to the best of our knowledge, is the first effort to do so.

Other efforts that can be considered as related to the proposed work include efforts in the unsupervised domain that have attempted the task of video prediction (predict a future frame given a video sequence). One of the first efforts in this regard, \cite{nitish_icml2015}, seeks to learn unsupervised video representations using Long Short-Term Memory units (LSTMs) \cite{hochreiter1997long} to predict future frames in videos. However, such efforts are fundamentally different from the objectives of this work, since they do not attempt to generate complete videos from scratch. We now present the proposed methodology.

\section{Sync-DRAW Methodology}
\label{sec_syncdraw}
Simply speaking, a video sequence can be perceived as comprising of objects undergoing/performing action(s) across multiple frames over a background. Hence, we primarily require our model to: (i) generate a video sequence; (ii) model individual objects in the video sequence; and (iii) capture the motion of the objects in the sequence.

To generate a video as a combination of several objects and to model their motion separately, it seems intuitive to generate the video incrementally with the network focusing and forming objects one at a time. A recurrent visual attention mechanism can prove to be an effective tool to \textit{attend} to relevant parts of a video sequence, thus allowing the network to learn latent representations of coherent parts of spatio-temporal data (objects). A video comprises of unique artifacts that are not associated with images (such as camera motion, e.g, zoom in/out), which have hitherto not been considered in image generation efforts. We hence propose a new read/write attention mechanism which operates at a frame level, and uses a grid of Gaussian filters with learnable variance and stride in each frame, which allows the network to automatically learn coherent regions of video sequences (inspired from \cite{draw,graves2014neural}). We also use a Recurrent VAE~(R-VAE), which has been used for image generation in ~\cite{draw} (but not video generation), to embed this frame-wise attention mechanism and generate a video over a set of time steps.

\begin{figure*}[t]
\centering
\includegraphics[width=0.8\textwidth]{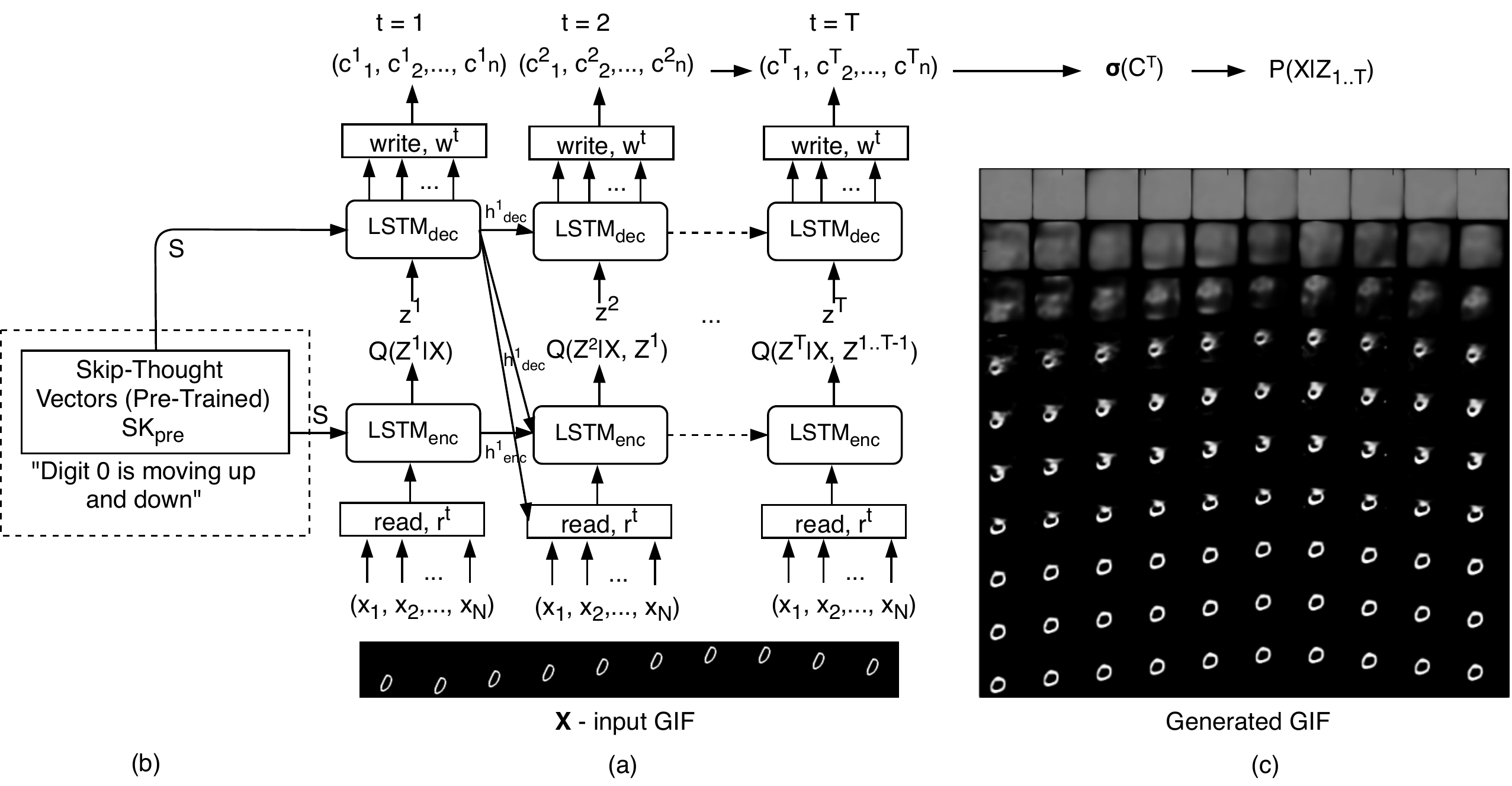}
\vspace{-10pt}
\caption{Sync-DRAW architecture. (a) The read-write mechanisms and the R-VAE; (b) Conditioning the R-VAE's decoder on text features extracted from captions for text-to-video generation; (c) Canvases generated by Sync-DRAW over time (each row is a video at a particular time step, last row is the final generated video).}
\vspace{-10pt}
\label{fig_syncdraw}
\end{figure*}

A trivial approach to extend the way R-VAE is used for image generation in ~\cite{draw} to video generation would be to have a separate R-VAE for generating each frame. However, such an approach ignores the temporal relationship between frames, and the number of parameters to be learnt across the VAEs may explode exponentially with increasing number of frames. Therefore, to learn the video data distribution efficienctly but effectively, the proposed Sync-DRAW uses a single R-VAE for the entire video (Figure \ref{fig_syncdraw}). This global R-VAE, firstly, allows us to model the temporal relationships between the generated video frames. Secondly, it also allows us to condition the generation of the video on captions to effectively perform text-to-video generation (this would not have been easy if we trivially extended the image R-VAE to videos). Moreover, while it is possible to define attention as a spatio-temporal cuboid (3-D grid of Gaussian filters) across frame blocks, we instead define a separate individual attention mechanism for each frame of a video, each of which acts simultaneously across all frames, so as to learn latent spatio-temporal representations of video data. We found this proposed strategy of using a global R-VAE with a local frame-wise attention mechanism to be key in the results that we obtained. This approach introduces smoothness in the characteristics of objects across the frames in comparison to 3-D spatio-temporal approaches~\cite{vondrick2016generating}, which seem to have jitters in the generations, as highlighted in the experiments section (Section \ref{sec_expts}).

\vspace{-5mm}
\subsection{Sync-DRAW Architecture}
\label{subsec_syncdraw_archi}

Let $ X = \{x_{1}, x_{2},\cdots, x_{N}\} $ be a video comprising of frames $x_{i}, i=1,\cdots, N$ where $N$ is the number of frames in the video. Let the dimensions for every frame be denoted by $A \times B$. We generate $X$ over $T$ timesteps, where at each timestep $t$, canvases for all frames are generated in synchronization. The Sync-DRAW architecture comprises of: (i) a \textit{read mechanism} which takes a region of interest (RoI) from each frame of the video; (ii) the \textit{R-VAE}, which is responsible to learn a latent distribution for the videos from these RoIs; and (iii) a \textit{write mechanism} which generates a canvas focusing on a RoI (which can be different from the reading RoI) for each frame. 
During training, at each time step, a RoI from each frame of the original video is extracted using the read mechanism and is fed to the R-VAE to learn the latent distribution, $Z$, of the video data. Then, a random sample from this distribution, $Z$, is used to generate the frames of the video using the frame-wise write mechanisms. At test time, the read mechanism and the encoder of the R-VAE are removed; and at each time step, a sample is drawn from the learned $Z$ distribution to generate a new video through the write mechanism. We now describe each of these components.



\subsubsection{Read Mechanism}
\label{subsubsec_syncdraw_read}
The read attention mechanism in Sync-DRAW is responsible for reading a region of interest (RoI) from each frame $x_{i}$ at every timestep $t$. As mentioned earlier, the RoI is defined by a $K \times K$ grid of Gaussian filters, where the grid's center, grid's stride and the Gaussian's variance (which is the same across all grid points) are all learned while training the network. Varying these learned parameters allows us to vary the RoI read at every time step. In particular, due to the separability of the 2-D Gaussian filter, the RoI patch is read by dynamically computing 1-D Gaussian filters $F_{i_p}^t$ and $F_{i_q}^t$, which correspond to a set of size $K$ (a user-defined parameter for the grid size) of $1 \times A$ and $1 \times B$-dimensional Gaussian filters respectively, for each frame $x_{i}$ at timestep $t$, with $i_p$ and $i_q$ being the two spatial dimensions of frame $x_i$. (For convenience, we will refer to $F_{i_p}^t$ and $F_{i_q}^t$ as being of sizes $K\times A$ and $K\times B$ respectively.) These filters are evaluated together on a $K \times K$ grid which defines the RoI of the frame to attend to.
In order to compute these filters, we learn a set of read attention parameters for each frame:
$\tilde{g}^t_{i_p},\tilde{g}^t_{i_q},\sigma_{i}^t,\tilde{\delta}^t_{i},\beta_{i}^t$ where $i\in \{1,\cdots,N\}$. $\tilde{g}^t_{i_p}$ and $\tilde{g}^t_{i_q}$ correspond to the grid center coordinates along the two frame dimensions respectively; and $\tilde{\delta}^t_{i}$ corresponds to the stride between the points in the grid. Evidently, we propose to learn a separate set of these parameters for each frame in the video. In order to ensure that these parameters lie within the frame dimensions, they are slightly modified using:
\[g^t_{i_p} = \dfrac{A + 1}{2} (\tilde{g}^t_{i_p} + 1); g^t_{i_q} = \dfrac{B + 1}{2} (\tilde{g}^t_{i_q} + 1) \]
\[ \delta^t_i = \dfrac{\max(A,B) - 1}{N - 1}\tilde{\delta}^t_{i} \]
These new parameters $g^t_{i_p}$ and $g^t_{i_q}$ are then used to calculate the centers for the horizontal and vertical filterbanks using:
\begin{align}
\mu^t_{i_{p_u}} = g^t_{i_p} + (u - N/2 - 0.5) \delta^t_i \\
\mu^t_{i_{q_v}} = g^t_{i_q} + (v - N/2 - 0.5) \delta^t_i
\end{align}
for $u,v \in \{1,\cdots,K\}$. $\sigma^t_{i}$ serves as the standard deviation for all filterbanks at timestep $t$ on $x_i$, the $i^{th}$ frame. $F_{i_p}^t$ and $F_{i_q}^t$ are then obtained finally as follows:
\begin{align}
F_{i_p}^t[u,a] = \dfrac{1}{Z_{i_p}} \exp \Big( - \frac{(a - \mu^t_{i_{p_u}})^2}{2(\sigma_i^t)^2} \Big)\\
F_{i_q}^t[v,b] = \dfrac{1}{Z_{i_q}} \exp \Big( - \frac{(b - \mu^t_{i_{q_v}})^2}{2(\sigma_i^t)^2} \Big)
\end{align}
where $a\in \{1,\cdots,A\}$, $b\in \{1,\cdots,B\}$, $u,v \in \{1,\cdots,K\}$ and $Z_{i_p}$, $Z_{i_q}$ are the normalization constants. 

The last parameter, $\beta_i^t$, allows the network to learn the temporal relationships between the patches read from the frames. It lets the model decide the level of importance to be given to each frame for generating the video at any time step. Before feeding to the R-VAE, the patch read from each frame of the input video is scaled by its respective $\beta_i^t, i\in \{1,\cdots,N\}$ as follows:
\begin{equation}
read(x_i) = \beta_i^t (F^t_{i_p}) x_i (F_{i_q}^t)^T
\end{equation}
where $read(x_i)$ is of size $K \times K$. This learned set of parameters, $\beta$, allows Sync-DRAW to selectively focus on one (or a subset) of frames of the video at a given timestep if required.\footnote{$\sigma$, $\tilde{\delta}$, and $\beta$ are defined in the logarithm scale to ensure that values are always positive.}

\subsubsection{R-VAE}
The R-VAE is responsible to generate the video sequence by sampling from the latent distribution learnt by the VAE at each time step. The core components of a standard R-VAE are the encoder LSTM, $LSTM_{enc}$ (which outputs $h_{enc}$), the latent representation, $Z$, and the decoder LSTM, $LSTM_{dec}$ (which outputs $h_{dec}$) \cite{rvae}. 
The R-VAE runs for $T$ time steps (user-defined) and generates a set of canvases $C^t = \{c_{1}^t, c_{2}^t,\cdots, c_{N}^t\}$ at every timestep $t$, where $\boldsymbol{\sigma}(C^t)$ is the video generated after $t$ timesteps. Since the R-VAE works at the global video level, the canvases for all the frames in the video are generated in synchronization, which by our observation, preserve the inter-frame temporal relationships. 
At every $t$, we define a new quantity, $\hat{X}^t$, which represents the video-level error, as follows:
\begin{equation}
\hat{X}^t = X - \boldsymbol{\sigma}(C^{t-1})
\end{equation} 
where $\boldsymbol{\sigma}$ is the sigmoid function, and $X$ is the input video. 
The LSTM-encoder at time $t$ is then processed as follows:
\begin{equation}
\label{eqn_htenc}
h_{enc}^{t} = LSTM_{enc} \left(h_{enc}^{t-1}, \left[R^t, h_{dec}^{t-1}\right]\right)
\end{equation}
where $R^t = \left[r^t_1, r^t_2,\cdots, r^t_N \right]$ with $r^t_i = [read(x_i), read(\hat{x}^t_i)]$. $h^t_{enc}$ is then used to estimate the parameters of latent variable $Z^{t}$, which characterizes the approximate posterior $Q\left(Z^t\vert h^t_{enc}\right)$ that captures the latent representation. This  is done using the reparameterization technique described in \cite{kingma2013auto}. A sample $z^t \sim Q\left(Z^t\vert h^t_{enc}\right)$ is then obtained and fed to the LSTM-decoder, $LSTM_{dec}$, producing $h_{dec}^{t}$: 
\begin{equation}
\label{eqn_htdec}
h^{t}_{dec} = LSTM_{dec} \left(h^{t-1}_{dec}, z^{t}\right)
\end{equation}
$h_{dec}^{t}$ is used to compute the write attention mechanism parameters as described below. We note that the read attention parameters at time $t$ are learned as a linear function of $h_{dec}^{t-1}$ and the input $X$, whose weights are learned during training in the read block. 


\subsubsection{Write Mechanism}
As part of the write mechanism, we need two things per frame: what to write, and where to write. The former is obtained using an attention window, $w^t$, which is computed as a linear function of $h^{t}_{dec}$, whose weights are learnt during training. The attention window is, once again, defined by an equivalently similar set of parameters as the read mechanism, which are now computed using $h^{t}_{dec}$ (as against using $h^{t-1}_{dec}$ and $X$ for the read parameters). Next, similar to the read mechanism, a set of filterbanks $\hat{F}_{i_p}^t$ and $\hat{F}_{i_q}^t$ are computed using these write attention parameters, defining the RoI to be attended to while writing into each frame. Finally, the canvas $c_i^t$ corresponding to every frame $i$ at time step $t$ is created as follows:
\begin{equation}
write(h_{dec}^t) = \frac{1}{\hat{\beta_i^t}} (\hat{F}_{i_p}^t)^T w_i^t (\hat{F}_{i_q}^t)
\end{equation}
\begin{equation}
\label{eqn_canvas}
c_i^t = c_i^{t-1} + write(h_{dec}^t)
\end{equation}

The canvasses are accumulated over timesteps T, denoted by $c^T$ and the final video generated is computed as $\sigma(c^T)$. By using sigmoid function over $c^T$, we can consider the output as the emission probabilities making it equivalent to $P(X|Z_1...T)$. In Figure~\ref{fig_syncdraw}, the final arrow is used to represent the connection between the technical implementation of the algorithm and the mathematical interpretation of the underlying model.

\subsubsection{Sync-DRAW with captions}
\label{sync-draw-with-captions}
To generate videos from captions, we use a pre-trained model of skip-thought vectors, $SK_{pre}$ ~\cite{kiros2015skip}, and pass the caption, $Y$, through this model to generate a latent representation for the caption, $S$. 

\begin{equation}
S = SK_{pre}(Y)
\end{equation}

We then condition $LSTM_{enc}$ and $LSTM_{dec}$ on $S$ ~\cite{sohn2015learning} to modify equations \ref{eqn_htenc} and \ref{eqn_htdec}, and generate corresponding hidden representations as:

\begin{eqnarray}
h_{enc}^{t} & = & LSTM_{enc} \left(h_{enc}^{t-1}, \left[R^t, h_{dec}^{t-1}, S\right]\right) \\
h^{t}_{dec} & = & LSTM_{dec} \left(h^{t-1}_{dec}, \left[z^{t}, S\right]\right)
\end{eqnarray}

At test time, when the R-VAE's encoder and read mechanisms are removed, a caption $Y$ is provided as input to the R-VAE's decoder to condition the generation on the given caption, $Y$. 
\subsubsection{Loss function}
The loss function for Sync-DRAW is composed of two types of losses (similar to a standard VAE \cite{kingma2013auto}), both of which are computed at the video level. The first is the reconstruction loss, $L_X$, computed as the binary pixel-wise cross-entropy loss between the original video $X$ and the generated video, $\boldsymbol{\sigma}(C^T)$. The second is the KL-divergence loss, $L_Z$, defined between a latent prior $P(Z^t)$ and $Q(Z^t\vert h^t_{enc}) \sim \mathcal{N}(\mu^t,(\sigma^t)^2)$ and summed over all $T$ timesteps. We assume prior $P(Z^t)$ as a standard normal distribution and thus, $L_Z$ is given by:
\begin{equation}
L_Z = \frac{1}{2} \Bigg( \sum\limits_{t=1}^T \mu_t^2 + \sigma_t^2 - \log \sigma_t^2 \Bigg) - T/2
\end{equation}
The final loss is the sum of the two losses, $L_X$ and $L_Z$.
\subsubsection{Testing}
During the testing phase, the encoder is removed from Sync-DRAW and due to the reparametrization trick of the VAE \cite{kingma2013auto}, $z$ is sampled from $\mathcal{N}\left(0,I\right)$, scaled and shifted by the learned $\mu^t$ and $(\sigma^t)^2$, to generate the final videos.
\begin{figure}[h]
\centering
\includegraphics[width=0.90\columnwidth]{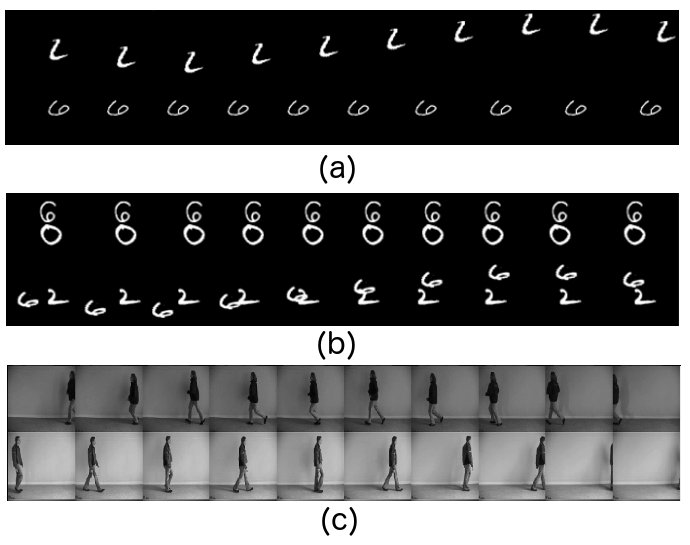}
\vspace{-15pt}
\caption{A few sample videos from each of the dataset used for evaluation. (a) Single-Digit Bouncing MNIST. (b) Two-Digit Bouncing MNIST. (c) KTH Dataset.}
\vspace{-10pt}
\label{dataset}
\end{figure}

\section{Experiments and Results}
\label{sec_expts}
We studied the performance of Sync-DRAW on the following datasets with varying complexity: (i) Single-Digit Bouncing MNIST (which has been used in similar earlier efforts \cite{kahou2015ratm,nitish_icml2015,xingjian2015convolutional}), (ii) Two-digit Bouncing MNIST; and (iii) KTH Human Action Dataset \cite{schuldt2004recognizing}. Figure~\ref{dataset} shows sample videos from these datasets. Considering this is the first work in text-to-video and there is no dataset yet for such a work, we chose these datasets since they provide different complexities, and also allow for adding captions\footnote{The codes, the captioned datasets and other relevant materials are available at \url{https://github.com/Singularity42/Sync-DRAW}}. We manually created text captions for each of these datasets (described later) to demonstrate Sync-DRAW's ability to generate videos from captions. 
We varied the number of timesteps $T$ and $K$ for read and write attention parameters across the experiments based on the size of the frame and complexity of the dataset (grayscale or RGB). Stochastic Gradient Descent with Adam \cite{kingma2014adam} was used for training with initial learning rate as $10^{-3}$, $\beta_1$ as $0.5$ and $\beta_2$ as $0.999$. Additionally, to avoid gradient from exploding, a threshold of $10.0$ was used.

\subsection{Baseline Methodology}
\label{baseline_method}
For performance comparison and to highlight the significance of having a separate set of attention parameters for each frame of the video, we designed another methodology to extend \cite{draw} to generate videos, by modeling attention parameters as spatio-temporal cuboids and adding a set of filterbanks $F^t_Z$ of size $K \times N$ to operate over the temporal dimension (in addition to the two filterbanks in the spatial dimensions). We use this methodology as our baseline for comparison. We tried on several settings of parameters and present the best results for the baseline methodology in this paper.

\begin{figure}[h]
\centering
\includegraphics[width=0.90\columnwidth]{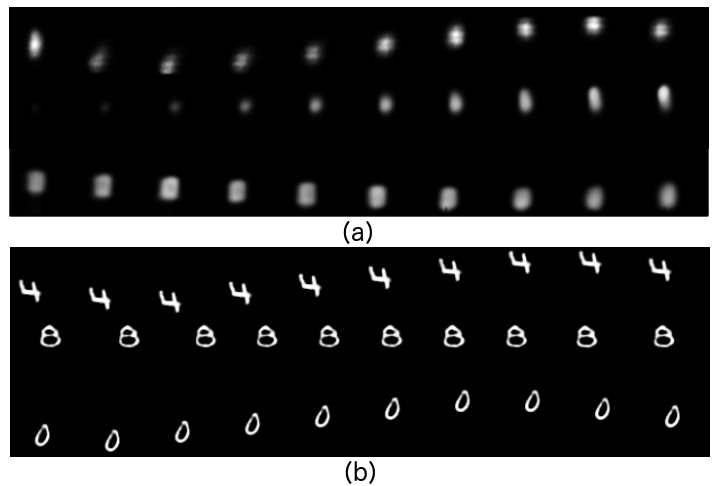}
\vspace{-15pt}
\caption{Single-Digit Bouncing MNIST results for (a) Baseline method and (b) Sync-DRAW}
\vspace{-15pt}
\label{single_digit_mnist_baseline_sync}
\end{figure}

\begin{figure}[h]
\centering
\includegraphics[width=0.90\columnwidth]{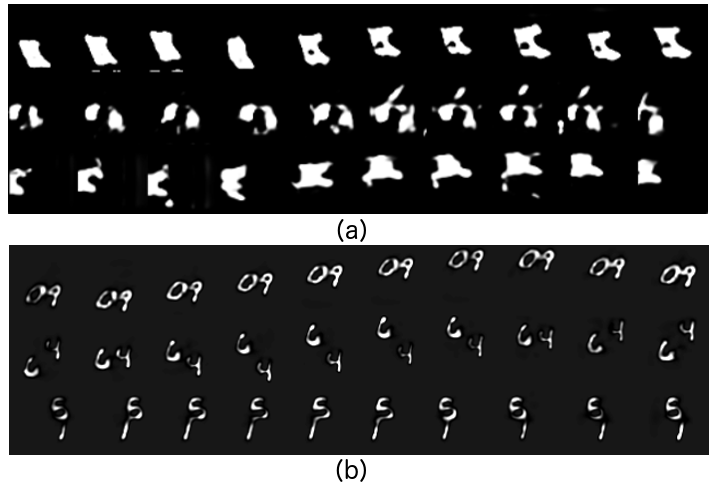}
\vspace{-15pt}
\caption{Two-Digit Bouncing MNIST results for (a) Baseline method and (b) Sync-DRAW}
\vspace{-20pt}
\label{two_digit_mnist_baseline_sync}
\end{figure}

\subsection{Results on Single-Digit Bouncing MNIST}
As in earlier work \cite{kahou2015ratm,nitish_icml2015,xingjian2015convolutional}, we generated the Single-Digit Bouncing MNIST dataset by having the MNIST handwritten digits move over time across the frames of the sequence (Figure~\ref{dataset}a). Each video contains $10$ frames each of size $64\times64$ with a single $28\times28$ digit either moving \textit{left and right} or \textit{up and down}. The initial position for the digit in each video is chosen randomly in order to increase variation among the samples. The training set contains $12,000$ such videos.
The results of Sync-DRAW on Single Digit MNIST are shown in Figures \ref{fig_syncdraw}(c) and \ref{single_digit_mnist_baseline_sync}(b). The figures illustrate the usefulness of the proposed methodology in gracefully generating the final video as a sequence of canvases. When compared to the baseline methodology results, as shown in Figure~\ref{single_digit_mnist_baseline_sync}(a), the quality of the generated digits is clearly superior with the digits having well-defined structure and boundaries. An interesting observation in Figure \ref{fig_syncdraw}(c) is that while each frame has its own attention mechanism in the proposed framework, it can be seen that the same region of the digit is being attended to at every timestep $t$, even though they are present at different pixel locations in each frame, thus validating the proposed approach to video generation. 

\begin{figure*}
\centering
\includegraphics[width=0.75\textwidth]{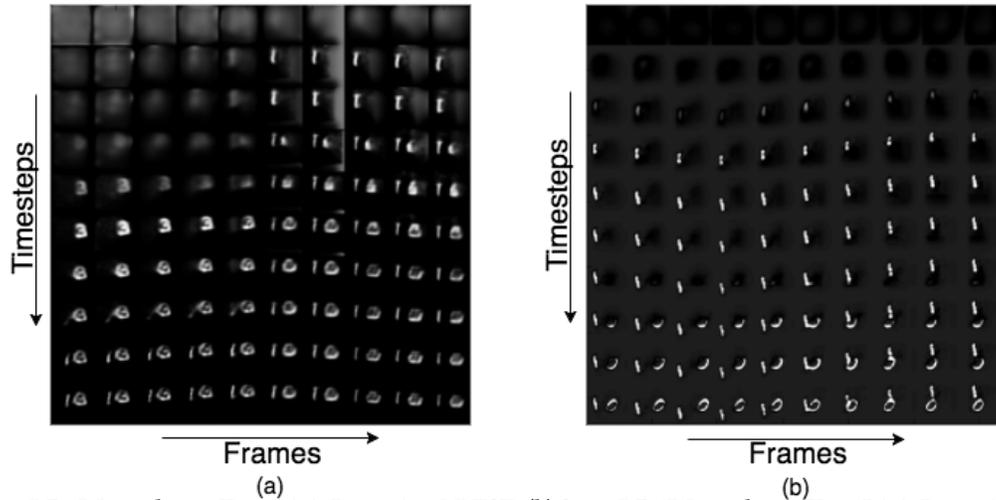}
\vspace{-15pt}
\caption{(a) Sync-DRAW results on Two-Digit Bouncing MNIST; (b) Sync-DRAW results on Two-Digit Bouncing MNIST videos when captions are included. Clearly, captions improve generation of `objectness' of the digits.}
\vspace{-10pt}
\label{results_2digitmnist}
\end{figure*}

\subsection{Results on Two-Digit Bouncing MNIST}
\label{subsec_results_2digit}
Extending the Bouncing MNIST dataset, we generated the two-digit Bouncing MNIST dataset where two digits move independent of one another, either \textit{up and down} or \textit{left and right}, as shown in Figure~\ref{dataset}(b). The dataset consists of a training set of $12,000$ videos, with each video containing 10 frames of size $64\times64$ with two $28\times28$ digits. In order to ensure variability, the initial positions for both the digits were chosen randomly. In case of an overlap, the intensities are added, clipping the sum if it goes beyond 1.  
The results of applying Sync-DRAW to this two-digit Bouncing MNIST dataset are shown in Figures \ref{two_digit_mnist_baseline_sync}(b) and \ref{results_2digitmnist}(a). Once again, the baseline method performs rather poorly as shown in Figure~\ref{two_digit_mnist_baseline_sync}(a). Figure \ref{results_2digitmnist}(a) also shows that Sync-DRAW attends to both the digits at every time step simultaneously. In the initial time steps, a single structure is formed which then breaks down to give the two digits in the subsequent time steps. We believe that this is the reason that though the generated digits have a well defined structure and a boundary, they are still not as clear as when compared to results on Single-Digit Bouncing MNIST. We infer that there is a need for a ``stimulus'' for attention mechanism to know that there are two digits that need to be attended to. This claim is substantiated in subsequent sections where we include alignment with captions in Sync-DRAW, giving the attention mechanism this required stimulus. 
\begin{figure}[h]
\centering
\includegraphics[width=\columnwidth]{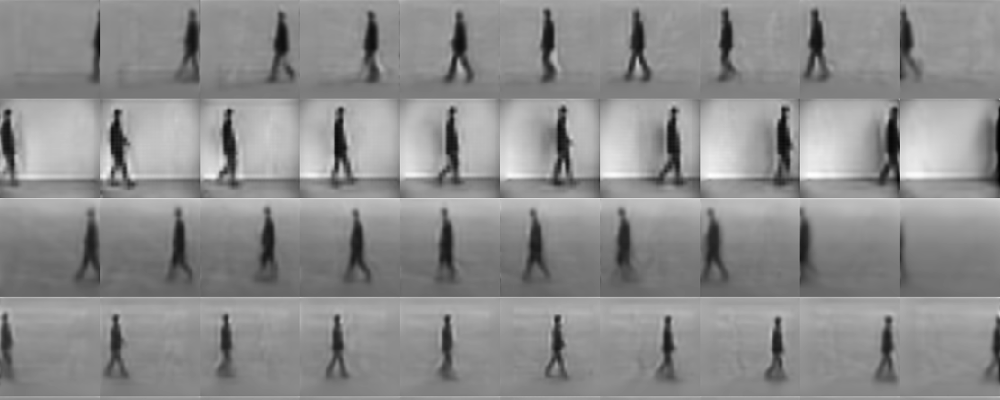}
\vspace{-15pt}
\caption{Videos generated by Sync-DRAW for KTH Human Action Dataset without captions.}
\label{kth_results}
\vspace{-15pt}
\end{figure}

\subsection{Results on KTH dataset}
\label{kth_no_caption}
We evaluated the performance of Sync-DRAW on the KTH Human Action Database~\cite{schuldt2004recognizing} to benchmark the approach on a more real dataset. We chose KTH dataset over other video datasets such as UCF as the videos from KTH were accompanied with metadata which helped us to create meaningful captions for the videos and experiment Sync-DRAW with captions on this dataset. The KTH dataset consists of over 2000 videos of 25 subjects performing six different actions - walking, running, jogging, hand-clapping, hand-waving and boxing. 
We resized the frames to $120 \times 120$ and performed generation for 10 frames. Given the complexity of the scene and increased frame size, we increased $K$ to $12$ and $T$ to $32$ to obtain videos with better resolution.

Figure~\ref{kth_results} shows some test videos generated for the KTH dataset. The clarity of the person in the generated scenes suggests that Sync-DRAW is able to perform equally well on a real-world dataset.

\begin{figure}[h]
\centering
\includegraphics[width=\columnwidth]{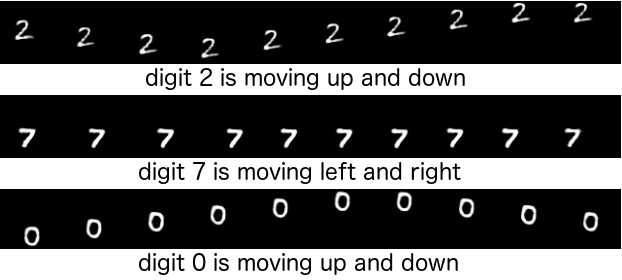}
\vspace{-15pt}
\caption{Sync-DRAW generates videos from just captions on the Single-Digit Bouncing MNIST: Results above were automatically generated by the trained model at test time.}
\vspace{-10pt}
\label{fig_single_digit_caption}
\end{figure}
\begin{figure}[h]
\centering
\includegraphics[width=\columnwidth]{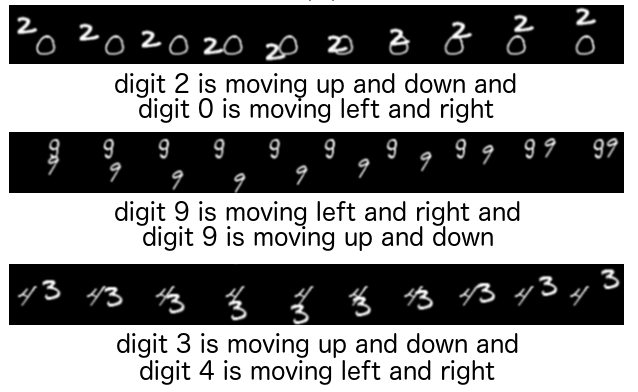}
\vspace{-15pt}
\caption{Sync-DRAW generates videos from just captions on the Two-Digit Bouncing MNIST: Results above were automatically generated by the trained model at test time.}
\vspace{-10pt}
\label{fig_two_digit_caption}
\end{figure}
\begin{figure}[h]
\centering
\includegraphics[width=\columnwidth]{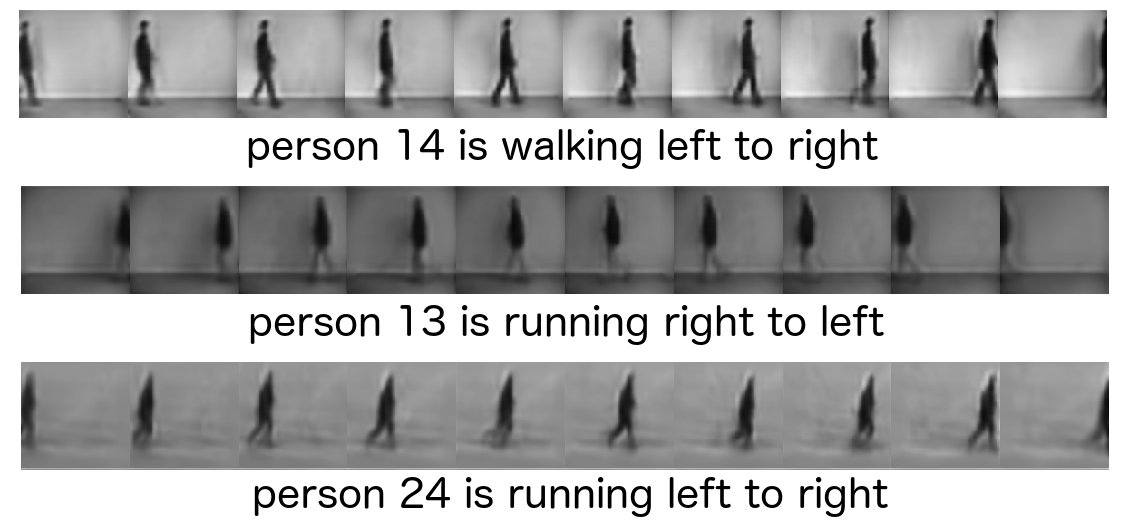}
\vspace{-15pt}
\caption{Sync-DRAW generates videos from just captions on KTH dataset.}
\vspace{-15pt}
\label{fig_kth_caption}
\end{figure}

\subsection{Sync-DRAW with captions}
\label{captions}
As discussed in Section~\ref{sync-draw-with-captions}, the proposed Sync-DRAW methodology can be used to generate videos from captions. 

\subsubsection{Single-Digit Bouncing MNIST} 
For every video in the bouncing MNIST dataset, a sentence caption describing the video was included in the dataset. For Single-Digit Bouncing MNIST, the concomitant caption was of the form \textit{'digit 0 is moving left and right'} or \textit{'digit 9 is moving up and down'}. 
Hence, for the single-digit dataset, we have 20 different combinations of captions. In order to challenge Sync-DRAW, we split our dataset in such a way that for all the digits, the training and the test sets contained different motions for the same digit, i.e. if a digit occurs with the motion involving \textit{up and down} in the training set, the caption for the same digit with the motion \textit{left and right} (which is not used for training) is used in the testing phase. Figure \ref{fig_single_digit_caption} shows some of the videos generated from captions present in the test set. It can be observed that even though the caption was not included in the training phase, Sync-DRAW is able to capture the implicit alignment between the caption, the digit and the movement fairly well.

\subsubsection{Two-Digit Bouncing MNIST} 
We conducted similar experiments for the Two-Digit Bouncing MNIST, where the captions included the respective motion information of both the digits, for example \textit{'digit 0 is moving up and down and digit 1 is moving left and right'}. Figure \ref{fig_two_digit_caption} shows the results on this dataset, which are fairly clear and good on visual inspection (we show quantitative results later in Section \ref{sec_quant_analysis}). Interestingly, in Figure~\ref{results_2digitmnist}(b), we notice that when Sync-DRAW is conditioned on captions, the quality of the digits is automatically enhanced, as compared to the results in Section \ref{subsec_results_2digit} (Figure~\ref{results_2digitmnist}(a)). These results give the indication that in the absence of captions (or additional stimuli), the attention mechanism in Sync-DRAW focuses on a small patch of a frame at a time, but possibly ignores the presence of different objects in the scene and visualizes the whole frame as one entity. However, by introducing captions, the attention mechanism receives the much needed ``stimulus'' to differentiate between the different objects and thereby cluster their generation, resulting in videos with better resolution. This is in concurrence with the very idea of an attention mechanism, which when guided by a stimulus, learns the spatio-temporal relationships in the video in a significantly better manner.

\subsubsection{KTH Dataset}
As mentioned in Section~\ref{kth_no_caption}, we were able to generate descriptive captions for the videos in the KTH dataset by using the metadata which included the person and the corresponding action. We carried out our experiments on videos for walking and running as it further helped to deterministically introduce the notion of direction. Some examples of the captions are \textit{'person 1 is walking right to left'} and \textit{'person 3 is running left to right'}. Figure~\ref{fig_kth_caption} shows some of the videos generated by Sync-DRAW using just the captions for KTH dataset. The generated videos clearly demonstrate Sync-DRAW's ability to learn the underlying representation of even real-world videos and create high quality videos from text.

\subsection{Quantitative analysis}
\label{sec_quant_analysis}
\subsubsection{Reconstruction Loss}
While quantitative analysis is difficult in image/video generation methods (due to lack of ground truth for the generations), we analyzed the quality of the videos generated by Sync-DRAW by computing the Negative Log Likelihood (NLL) of the generated samples (as recommended by Theis et al. in ~\cite{theis2015note} for evaluating generative models). We compared the NLL values at convergence against the baseline approach, and a setup where, instead of a hierarchical approach, we allowed both the spatial and temporal relationships to be learned at the global level (called Global in table below). The results are presented in Table~\ref{LLTable}. From the losses, we can clearly infer that Sync-DRAW performs significantly better than the other two approaches.

\begin{table}[h]
\small
\centering
\begin{tabular}{|l|l|l|l|l|}
\hline
\hline
Experiments & \begin{tabular}[c]{@{}l@{}}One Digit\\ MNIST\end{tabular} & \begin{tabular}[c]{@{}l@{}}One Digit\\ MNIST*\end{tabular} & \begin{tabular}[c]{@{}l@{}}Two Digit\\ MNIST\end{tabular} & \begin{tabular}[c]{@{}l@{}}Two Digit\\ MNIST*\end{tabular} \\ \hline \hline
Sync-DRAW & 340.39 & 327.11 & 639.71 & 524.41 \\ \hline
Global & 500.01 & 512.46 & 899.06 & 860.54 \\ \hline
Baseline & 3561.23 & 3478.26 & 5240.65 & 5167.94 \\ \hline \hline
\end{tabular}
\caption{Negative Log Likelihoods at convergence for Sync-DRAW, Baseline and Global methods. * = with captions (Lower is better).}
\vspace{-20pt}
\label{LLTable}
\end{table}
\subsubsection{Psychophysical Analysis}
In order to quantify the visual quality of the generated videos, we further performed a psychophysical analysis where we asked a group of human subjects to rate the videos generated by Sync-DRAW for different datasets. $37$ subjects (each ignorant of this work) were given a set of $10$ generated videos from each experiment we conducted with Sync-DRAW, and were asked to rate the perceived visual quality of a video on a scale of $1-10$ (with $10$ being the highest score, and also corresponded to the score for original videos from the respective datasets shown to the subjects). The statistics of this analysis are presented in Tables~\ref{PP without captions} and \ref{PP with caption}. It can be observed that the average rating for Single-Digit Bouncing MNIST (with and without captions), Two-Digit Bouncing MNIST (with captions) and KTH (with captions) is very high which correlates well with the qualitative results. The scores for Two-Digit Bouncing MNIST and KTH without captions were slightly lower (although not unacceptable) due to the reduction in clarity. Generations on these datasets, however, improved with the availability of a stimulus in the form of captions (as  explained before in Figure~\ref{results_2digitmnist}).


\begin{table}[]
\centering
\begin{tabular}{|l|l|l|l|}
\hline
\hline
Datasets                                                            & \begin{tabular}[c]{@{}l@{}}One Digit \\ MNIST\end{tabular} & \begin{tabular}[c]{@{}l@{}}Two Digit\\  MNIST\end{tabular} & KTH Dataset     \\ \hline \hline
\begin{tabular}[c]{@{}l@{}}Avg. Video\\ Quality Score\end{tabular}&9.37 $\pm$ 0.63&7.86 $\pm$ 0.75&7.27 $\pm$ 0.80 \\ \hline \hline
\end{tabular}
\caption{Average score $\pm$ Standard Deviation given to the videos generated by Sync-DRAW without captions for different datasets. Baseline: low score ($2.7 \pm 0.53$) for all datasets.}
\label{PP without captions}
\vspace{-15pt}
\end{table}


\begin{table}[]
\centering
\begin{tabular}{|l|l|l|l|}
\hline
\hline
Datasets                                                            & \begin{tabular}[c]{@{}l@{}}One Digit \\ MNIST\end{tabular} & \begin{tabular}[c]{@{}l@{}}Two Digit\\  MNIST\end{tabular} & KTH Dataset     \\ \hline \hline
\begin{tabular}[c]{@{}l@{}} Avg. Video\\ Quality Score\end{tabular}&9.43 $\pm$ 0.60&8.97 $\pm$ 0.83&9.10 $\pm$ 0.77 \\ \hline \hline
\end{tabular}
\caption{Average score $\pm$ Std Deviation given to videos generated by Sync-DRAW with captions for different datasets. }
\label{PP with caption}
\vspace{-15pt}
\end{table}


\begin{figure}[h]
\centering
\includegraphics[width=\columnwidth]{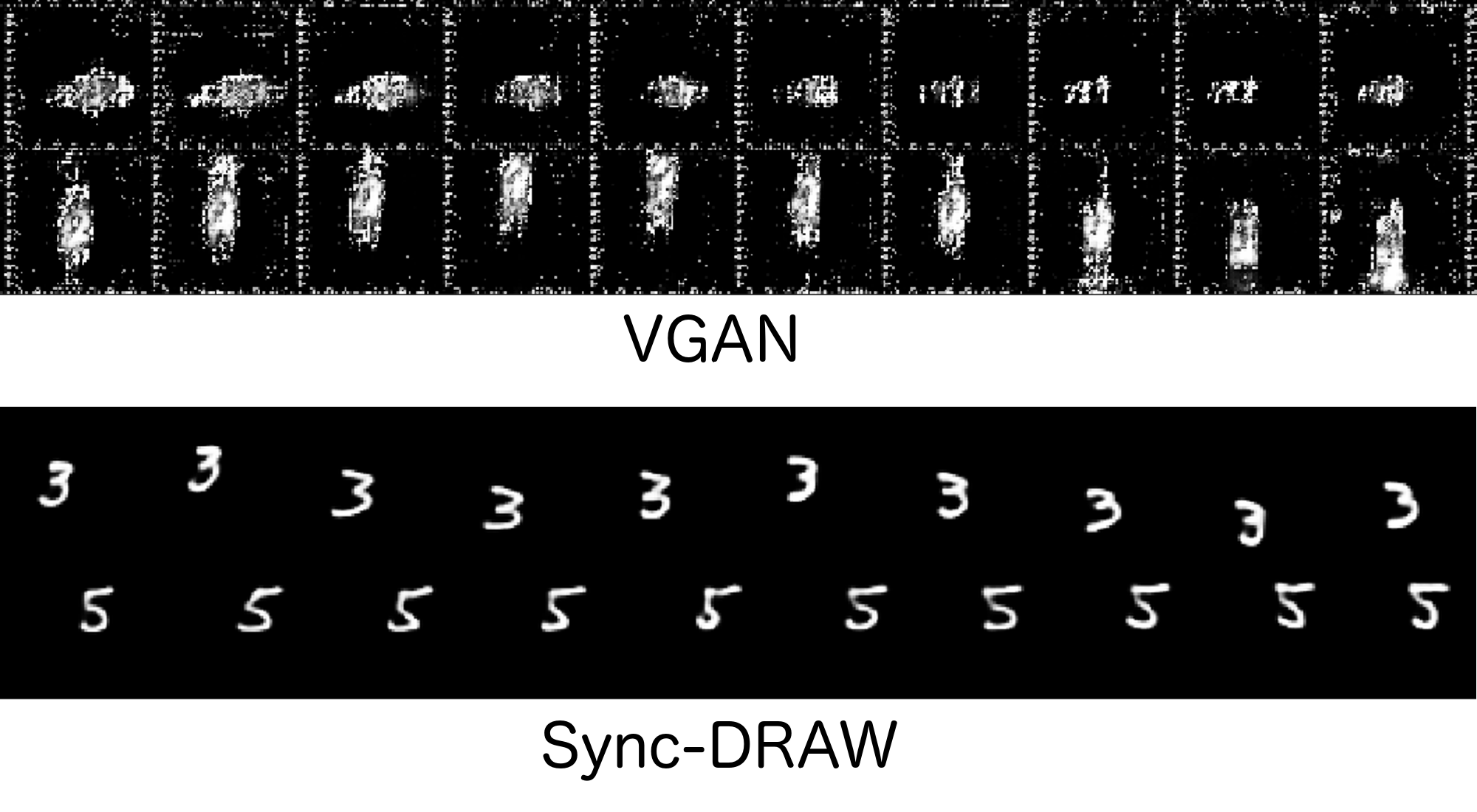}
\vspace{-15pt}
\caption{Qualitative comparison of video generation on MNIST dataset between VGAN and Sync-DRAW. First 10 frames generated by both approaches are shown for brevity.}
\vspace{-10pt}
\label{fig_vgan_syncdraw_mnist}
\end{figure}

\begin{figure}[h]
\centering
\includegraphics[width=\columnwidth]{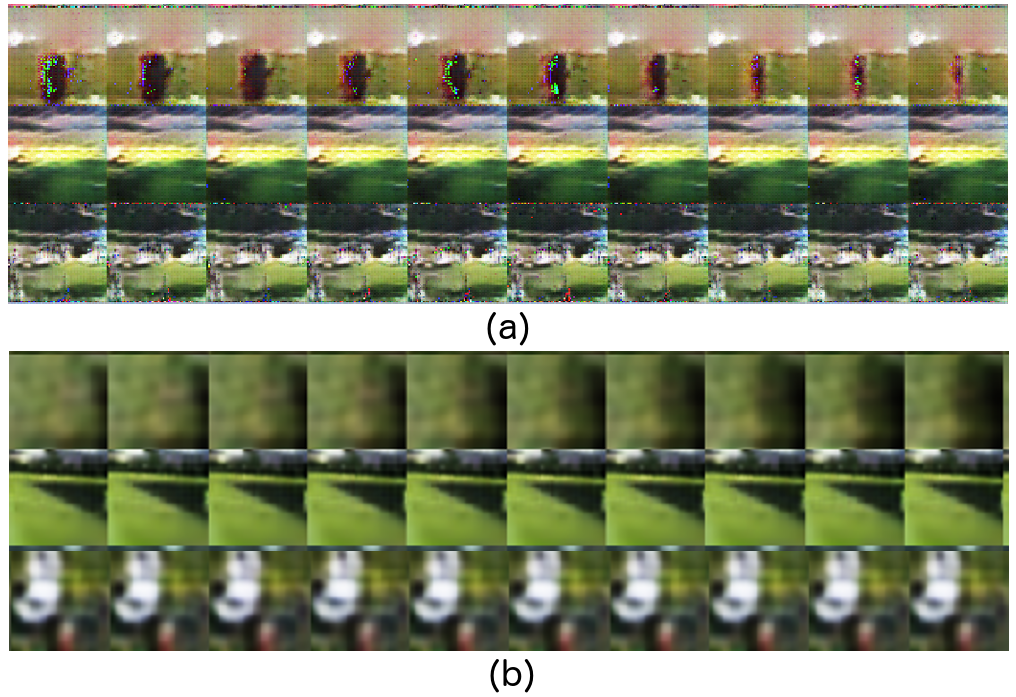}
\vspace{-15pt}
\caption{Qualitative comparison of video generation on dataset in ~\cite{vondrick2016generating} between VGAN and Sync-DRAW. (a) depicts the videos generated by VGAN model we trained. (b) depicts the videos generated by Sync-DRAW. The first 10 frames generated by both approaches are shown here for brevity.}
\vspace{-10pt}
\label{fig_vgan_syncdraw_mit}
\end{figure}

\begin{figure}[h]
\centering
\includegraphics[width=\columnwidth]{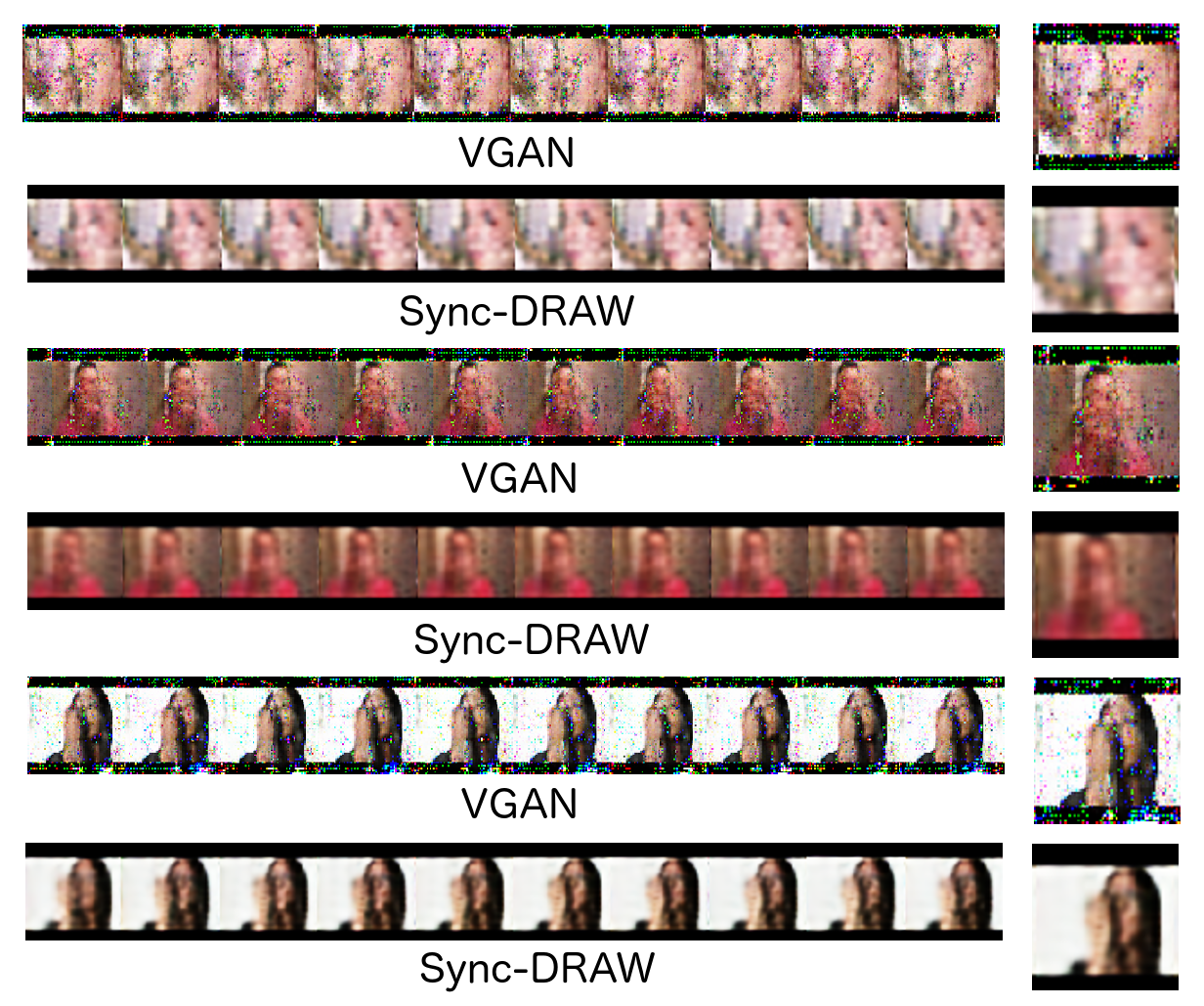}
\vspace{-15pt}
\caption{Qualitative comparison of video generation on UCF101 dataset between VGAN and Sync-DRAW. The first 10 frames generated by both approaches are shown here for brevity. On the right, a sample frame of each video is shown magnified for better comparison.}
\vspace{-10pt}
\label{fig_vgan_syncdraw_ucf}
\end{figure}
\subsection{Qualitative comparison with VGAN \cite{vondrick2016generating}}
\label{sec_vgan_compare}
To compare our results with the recently proposed VGAN method in \cite{vondrick2016generating}, which is the only earlier effort for video generation from scratch, we ran a comprehensive set of experiments. We first studied the performance of VGAN's~\cite{vondrick2016generating} code on the Bouncing MNIST dataset. In order to align with VGAN's requirements, we modified the Bouncing MNIST dataset to comprise of videos containing $32$ frames in RGB, and then trained VGAN on this modified Bouncing MNIST dataset. To ensure an equitable comparison, we ran Sync-DRAW on the same dataset as well (note that our earlier results had different settings on this dataset). The results of the experiments are shown in Figure~\ref{fig_vgan_syncdraw_mnist}. Next, we ran Sync-DRAW on the dataset made available in VGAN's work~\cite{vondrick2016generating}. We changed Sync-DRAW's parameters to support generation of videos containing $32$ frames. The results can be seen in Figure~\ref{fig_vgan_syncdraw_mit}. Lastly, in order to have a fair comparison, we ran both the codes on a dataset that they were never run on before, the UCF-101 dataset~\cite{soomro2012ucf101}. Each video in the training dataset from UCF-101 contained $32$ colored frames. \footnote{VGAN \cite{vondrick2016generating} uses UCF-101 to show their network's capability to learn representations for action classification but has shown no generation result on UCF-101} These results can be seen in Figure~\ref{fig_vgan_syncdraw_ucf}.

From our experiments, we notice that there are significant differences between Sync-DRAW and VGAN results. While VGAN gives an overall semblance of sharper generations, on closer look, the generations are noisy with poor objectness. (While \cite{vondrick2016generating} reported results that look better, we found this to be true on closer observation for many results there too). On the other hand, Sync-DRAW's generations have a smoothness to them and are fairly noise-free in comparison, although they are a bit blurry. The results of VGAN on Bouncing MNIST highlight this issue of poor objectness further. Similar observations can also be made on the results generated from training on VGAN~\cite{vondrick2016generating}'s own dataset. 
The results on UCF-101 serve as a fair comparison between the two methods because of the lack of any bias in the dataset used to train the two models. We also conducted a psychophysical analysis on the videos generated by VGAN, and found the scores on VGAN to be comparable to that of Sync-DRAW. Sharpening the generations of Sync-DRAW on real-world datasets, while maintaining its overall smoothness, is an important direction of our future work.

\section{Conclusion and Future Work}
This paper presents Sync-DRAW, a new approach to automatically generate videos using a Variational Auto-Encoder and a Recurrent Attention Mechanism combined together. We demonstrate Sync-DRAW's capability to generate increasingly complex videos starting from Single-Digit Bounding MNIST to more complex videos from KTH and UCF101 datasets. We show that our approach gives adequate focus to the objects via the attention mechanism and generates videos that maintain the structural integrity of objects. We also demonstrated Sync-DRAW's ability to generate videos using just captions, which is the first of its kind. As evident from the results, our approach is highly promising in the domain of video generation. Our ongoing/future efforts are aimed at improving the quality of the generated videos by reducing blurriness and generating videos from even more complex captions. 

\bibliographystyle{ACM-Reference-Format}
\balance
\bibliography{sigproc}

\end{document}